\documentclass[10pt,journal]{IEEEtran}
\usepackage{amsmath,amsfonts}
\usepackage{algorithmic}
\usepackage{algorithm}
\usepackage{array}
\usepackage[caption=false,font=normalsize,labelfont=sf,textfont=sf]{subfig}
\usepackage{textcomp}
\usepackage{stfloats}
\usepackage{url}
\usepackage{verbatim}
\usepackage{graphicx}
\usepackage{cite}
\usepackage{amsmath,amsfonts,bm}

\def\eqref#1{equation~\ref{#1}}

\def\1{\bm{1}}

\def\vs{{\bm{s}}}

\DeclareMathAlphabet{\mathsfit}{\encodingdefault}{\sfdefault}{m}{sl}
\SetMathAlphabet{\mathsfit}{bold}{\encodingdefault}{\sfdefault}{bx}{n}

\newcommand{\Ls}{\mathcal{L}}

\usepackage[dvipsnames]{xcolor}
\definecolor{Gray}{gray}{0.9}
\definecolor{mygreen}{rgb}{0.0, 0.5, 0.0}
\definecolor{myred}{rgb}{0.8, 0.25, 0.33}
\definecolor{myblue}{rgb}{0.19, 0.55, 0.91}
\definecolor{uclablue}{rgb}{0.15, 0.45, 0.68}
\definecolor{ucladblue}{rgb}{0.0, 0.33, 0.53}
\definecolor{ucladdblue}{rgb}{0.0, 0.23, 0.36}
\definecolor{uclagold}{rgb}{1.0, 0.82, 0.0}
\definecolor{ucladgold}{rgb}{1.0, 0.78, 0.17}
\definecolor{ucladdgold}{rgb}{1.0, 0.72, 0.11}
\definecolor{boxgreen}{rgb}{0.02, 0.66, 0.02}
\definecolor{boxred}{rgb}{0.66, 0.1, 0.1}
\definecolor{boxblue}{rgb}{0.01, 0.01, 0.73}

\usepackage{etoc}
\usepackage{lipsum}
\usepackage{wrapfig}
\usepackage{multicol}
\usepackage{mdframed}

\usepackage{mathtools}
\usepackage{amsthm}

\usepackage[utf8]{inputenc}   %
\usepackage[T1]{fontenc}      %
\usepackage{url}              %
\usepackage{amsfonts}         %
\usepackage{nicefrac}         %
\usepackage{microtype}
\usepackage{graphicx}
\usepackage{amsmath,amssymb,mathbbol}
\usepackage{acronym}
\usepackage{enumitem}
\usepackage{balance}
\usepackage{xspace}
\usepackage{setspace}
\usepackage[skip=3pt,font=small]{caption}
\usepackage{booktabs,tabularx,colortbl,multirow,array,makecell}
\usepackage{pifont}
\usepackage{xr-hyper}
\usepackage{tcolorbox}
\tcbuselibrary{breakable}
\usepackage{newtxmath}
\usepackage{dsfont}

\usepackage[pagebackref,breaklinks,citecolor=ucladblue,colorlinks=true,linkcolor=ucladblue,bookmarks=false]{hyperref}
\usepackage[capitalise]{cleveref}

\usepackage{siunitx}   %
\usepackage{soul}   %
\usepackage{listings}
\definecolor{codegreen}{rgb}{0,0.6,0}
\definecolor{codegray}{rgb}{0.5,0.5,0.5}
\definecolor{codepurple}{rgb}{0.58,0,0.82}
\definecolor{backcolour}{rgb}{0.95,0.95,0.92}

\lstdefinestyle{mystyle}{
    backgroundcolor=\color{backcolour},   
    commentstyle=\color{codegreen},
    keywordstyle=\color{magenta},
    numberstyle=\tiny\color{codegray},
    stringstyle=\color{codepurple},
    basicstyle=\ttfamily\footnotesize,
    breakatwhitespace=false,         
    breaklines=true,                 
    captionpos=b,                    
    keepspaces=true,                 
    numbers=left,                    
    numbersep=5pt,                  
    showspaces=false,                
    showstringspaces=false,
    showtabs=false,                  
    tabsize=2
}

\makeatletter
\DeclareRobustCommand\onedot{\futurelet\@let@token\@onedot}
\def\@onedot{\ifx\@let@token.\else.\null\fi\xspace}
\def\eg{\emph{e.g}\onedot} 

\def\ie{\emph{i.e}\onedot}

\def\etc{\emph{etc}\onedot} 
\def\vs{\emph{vs}\onedot}

\makeatother

\makeatletter
\renewcommand{\paragraph}[1]{%
  \vspace{1.5ex plus 0.5ex minus 0.2ex}%
  \noindent\normalfont\normalsize\textbf{#1}\hspace{1em}%
}
\makeatother

\crefname{algorithm}{Alg.}{Algs.}
\Crefname{algocf}{Algorithm}{Algorithms}
\crefname{section}{Sec.}{Secs.}
\Crefname{section}{Section}{Sections}
\crefname{table}{Tab.}{Tabs.}
\Crefname{table}{Table}{Tables}
\crefname{figure}{Fig.}{Fig.}
\Crefname{figure}{Figure}{Figure}

\definecolor{gblue}{HTML}{4285F4}
\definecolor{gred}{HTML}{DB4437}
\definecolor{ggreen}{HTML}{0F9D58}

\definecolor{mygray}{gray}{.92}

\newcommand{\supp}{\textit{Appendix}\xspace}

\newcommand{\sota}{state-of-the-art\xspace}

\newcommand{\cmark}{\ding{51}}
\newcommand{\xmark}{\ding{55}}

\newcommand{\model}{\textsc{LEO-VL}\xspace}

\acrodef{llm}[LLM]{large language model}
\acrodef{vla}[VLA]{vision-language-action}
\acrodef{vl}[VL]{vision-language}
\acrodef{qa}[QA]{question answering}
\acrodef{2dvl}[2D-VL]{2D vision-language}
\acrodef{3dvl}[3D-VL]{3D vision-language}
\acrodef{ocot}[O-CoT]{Object-centric Chain-of-Thought}
\acrodef{cot}[CoT]{Chain-of-Thought}
\acrodef{vlm}[VLM]{vision-language model}
\acrodef{lvlm}[LVLM]{large vision-language model}
\acrodef{rl}[RL]{reinforcement learning}
\acrodef{sft}[SFT]{supervised fine-tuning}
\acrodef{rlhf}[RLHF]{reinforcement learning from human feedback}
\acrodef{cfg}[CFG]{condensed feature grid}
\acrodef{rope}[RoPE]{rotary position embedding}
\acrodef{nlp}[NLP]{natural language processing}
\acrodef{dpo}[DPO]{direct preference optimization}
\acrodef{grpo}[GRPO]{group relative policy optimization}
\acrodef{iou}[IoU]{Intersection over Union}
\acrodef{id}[ID]{in-domain}
\acrodef{ood}[OOD]{out-of-domain}
\acrodef{nll}[NLL]{negative log likelihood}

\definecolor{scannet}{rgb}{0.7,0.2,0.1}
\definecolor{3rscan}{rgb}{0.1,0.6,0.2}
\definecolor{multiscan}{rgb}{0.1,0.3,0.6}
\definecolor{arkitscenes}{rgb}{0.5,0.1,0.5}

\definecolor{line1}{rgb}{0.99,0.96,0.96}
\definecolor{line2}{rgb}{0.96,0.99,0.94}
\definecolor{line3}{rgb}{0.92,0.96,1.0}

\begin{document}

\title{LEO-VL: Efficient Scene Representation for \\ Scalable 3D Vision-Language Learning}

\author{Jiangyong Huang,
Xiaojian Ma,
Xiongkun Linghu,
Junchao He,\\%
Qing Li,
Song-Chun Zhu,
Yixin Chen,
Baoxiong Jia,
Siyuan Huang\\%
\vspace{1em}%

\thanks{Jiangyong Huang is with the Institute for Artificial Intelligence, Peking University, Beijing, China, and also with the State Key Laboratory of General Artificial Intelligence, BIGAI, Beijing, China (e-mail: huangjiangyong@pku.edu.cn).}
\thanks{Xiaojian Ma, Xiongkun Linghu, Qing Li, Yixin Chen, Baoxiong Jia, and Siyuan Huang are with the State Key Laboratory of General Artificial Intelligence, BIGAI, Beijing, China (e-mail: baoxiongjia@g.ucla.edu; huangsiyuan@ucla.edu).}
\thanks{Junchao He is with the School of Artificial Intelligence, Beijing University of Posts and Telecommunications, Beijing, China, and also with the State Key Laboratory of General Artificial Intelligence, BIGAI, Beijing, China.}
\thanks{Song-Chun Zhu is with the State Key Laboratory of General Artificial Intelligence, BIGAI, Beijing, China, also with the Institute for Artificial Intelligence, Peking University, Beijing, China, and also with the Department of Automation, Tsinghua University, Beijing, China.}
}

\setcounter{secnumdepth}{3}
\renewcommand{\arraystretch}{1.1}

\makeatletter
\patchcmd{\@maketitle}
  {\addvspace{0.5\baselineskip}\egroup}
  {\addvspace{0.5\baselineskip}\egroup
  \begin{center}
     \begin{minipage}{0.92\linewidth}
     \includegraphics[width=\linewidth]{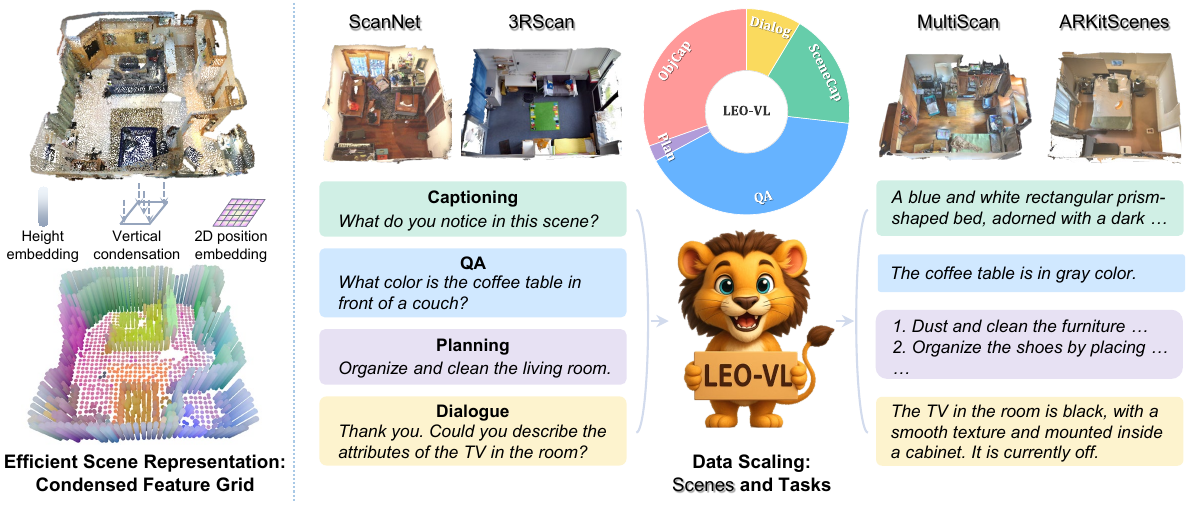}
     \setcounter{figure}{0}
     \captionsetup{hypcap=false}
     \vspace{-1em}
     \captionof{figure}{\textbf{\model overview.} \model features an efficient scene representation with strong perception capability and low representation costs, unlocking the scalability of \ac{3dvl} learning across diverse scene domains (\eg, ScanNet, 3RScan) and tasks such as captioning, dialogue, \etc.}
     \captionsetup{hypcap=true}
     \label{fig:teaser}
   \end{minipage}
  \end{center}\vspace{-1em}}
  {}
  {}
\makeatother

\maketitle

\begin{abstract}
Developing \acfp{vlm} capable of understanding 3D scenes has been a longstanding research goal. Despite recent progress, 3D \acp{vlm} still struggle with spatial reasoning and robustness. We identify three key obstacles hindering their progress: (1) scene representation is constrained by a capacity-efficiency trade-off, which impedes scalable learning; (2) training data lacks a comprehensive scheme, with limited diversity across tasks and scene domains; and (3) models exhibit robustness deficiencies and lack effective post-training. To address these challenges, we first propose \acf{cfg}, an efficient scene representation that significantly reduces token overhead while preserving strong perceptual capacity. Building on \ac{cfg}, we introduce \model, a 3D \ac{vlm} trained on over 700k \acf{3dvl} data spanning four real-world indoor domains and five tasks such as captioning and dialogue. To further improve robustness, we propose SceneDPO, a novel post-training objective that incorporates contrastive signals across both answers and scenes. \model achieves \sota performance on various \ac{3dvl} benchmarks, such as SQA3D, Beacon3D, and Scan2Cap. Extensive analyses highlight the efficiency of \ac{cfg} and provide key insights such as the importance of task and scene diversity, the priority of data quality for effective scaling, and the advantages of SceneDPO.
\end{abstract}

\section{Introduction}
\label{sec:intro}

\IEEEPARstart{3}{D} scene understanding represents a cornerstone of human intelligence. In pursuit of replicating this ability, a primary goal within the \acf{3dvl} community has been the development of 3D \acfp{vlm} that can understand 3D environments and communicate with humans through natural language \cite{huang2024embodied,yang2025thinking,song2025robospatial,ma2024llms}. Despite advancement in foundational \acp{vlm} \cite{openai2023gpt4,team2023gemini,wang2024qwen2,chen2024internvl,li2024llava} and progress in specific \ac{3dvl} tasks \cite{zhu20233d,zhu2024unifying,jia2024sceneverse,luo2023scalable,xu2024pointllm,huang2024chatscene,linghu2024multi,huang2024embodied,yang2024llm}, current 3D \acp{vlm} still fall short of comprehensive scene understanding, particularly regarding spatial understanding and complex reasoning in 3D scenes. We identify three critical obstacles that hinder the advancement of 3D \acp{vlm}: (1) a fundamental bottleneck in scene representation, dictated by a trade-off between capacity and efficiency, which impedes scalable learning; (2) the absence of a comprehensive data scheme that encompasses diverse tasks and scene domains; and (3) the lack of a robust post-training objective to enhance downstream capability.

Existing scene representations for 3D \acp{vlm} generally follow two paradigms. The first adopts direct 3D modality inputs, such as point clouds, which require complex pre-processing pipelines, \eg, 3D reconstruction, and instance segmentation. This approach is often constrained by the inherent difficulty of 3D perception and the scarcity of high-quality \ac{3dvl} data. The second paradigm employs 2D images or videos as input to leverage the mature perception power of 2D visual encoders. Despite superior performance, this paradigm incurs substantial token overhead and severely restricts data scalability \cite{kong2025token}. Additionally, both paradigms typically rely on rudimentary position embeddings that lack inductive bias for 3D awareness, which can be difficult to learn and may fall short in capturing nuanced 3D structures.

To overcome these limitations, we propose the \acf{cfg}, a novel scene representation that drastically improves efficiency while preserving strong perception capability. Given multi-view RGB-D inputs, we employ 2D \acp{vlm} to extract visual features, which are then back-projected into 3D voxels. We encode vertical height information using \ac{rope} \cite{su2024roformer} and condense voxels within each pillar region into efficient grid tokens. We further inject horizontal position embeddings via 2D Fourier features \cite{tancik2020fourier}. This design reduces the scene token overhead to $33\%$ while maintaining global 3D structural integrity. Based on the \ac{cfg} representation, we build \model, which integrates \ac{cfg} tokens with a \acf{llm} for auto-regressive language modeling to handle various \ac{3dvl} tasks.

As representation efficiency unlocks scalable \ac{3dvl} learning, we construct a comprehensive \ac{3dvl} dataset comprising over 700k high-quality samples, spanning four real-world indoor domains \cite{dai2017scannet,wald2019rio,mao2022multiscan,baruch2021arkitscenes} and five \ac{3dvl} tasks, covering captioning, \acf{qa}, planning, and dialogue. Our construction incorporates both existing data sources and newly generated samples, prioritizing data quality and diversity over raw scale to ensure consistent scaling effects and prevent model performance degradation.

Furthermore, we argue that standard \ac{sft} is insufficient to develop robust 3D \acp{vlm} given the prevailing overfitting issues in current models \cite{deng2024can,huang2025unveiling}. To address this, we propose SceneDPO, a post-training objective that adapts \ac{dpo} for \ac{3dvl} tasks. SceneDPO contrasts positive (\ie, correct) answers against hard negative answers, and encourages the model to exploit scene context by discouraging positive answers when conditioned on irrelevant scene context. We also incorporate a \acf{nll} loss to ensure optimization stability and prevent model degradation. The overall SceneDPO objective is dedicated to cultivating robust capability in 3D scenes while suppressing overfitting issues.

Our comprehensive evaluations demonstrate that \model achieves \sota performance with significantly higher efficiency across various \ac{3dvl} benchmarks, such as SQA3D,  Beacon3D, and Scan2Cap. Our model analyses highlight the effectiveness and efficiency of the \ac{cfg} representation, while our data analyses underscore the importance of task diversity (\eg, captioning and dialogue), and scene diversity for cross-domain performance. Notably, we observe that naive scaling with low-quality QA data degrades performance, whereas our data curation principle yields consistent scaling effects. For post-training, we demonstrate that SceneDPO exhibits superior \acf{id} and \acf{ood} performance compared to \ac{sft} and \ac{grpo} \cite{shao2024deepseekmath}, alongside effective optimization trends during post-training.

In summary, our contributions are as follows:
\begin{enumerate}[nolistsep,noitemsep,leftmargin=*]
    \item We introduce \model, a 3D \ac{vlm} equipped with the \acf{cfg} representation, featuring versatile \ac{3dvl} capabilities and significantly improved efficiency, which facilitates scalable \ac{3dvl} learning.
    \item We construct a comprehensive \ac{3dvl} dataset of over 700k samples across four real-world scene domains and five tasks, establishing a principle that prioritizes diversity and quality to ensure effective scaling for \ac{3dvl} learning.
    \item We propose SceneDPO, a \ac{3dvl} post-training objective that improves model robustness and mitigates overfitting.
    \item Our comprehensive evaluations demonstrate \model's \sota performance across various \ac{3dvl} benchmarks. And our extensive analyses provide valuable insights into model design, data scaling, and post-training.
\end{enumerate}

\section{Related Work}
\label{sec:related_work}

\paragraph{3D Vision-Language Models.} Most early works in \ac{3dvl} understanding develop models based on 3D representations such as point clouds \cite{qi2017pointnet,qi2017pointnet++,phan2018dgcnn,li2018pointcnn,wu2019pointconv,qi2019deep,zhao2021point,huang2021spatio,yu2022pointbert} and voxels \cite{graham2015sparse,maturana2015voxnet,riegler2017octnet,tatarchenko2017octree,graham20183d,dai20183dmv,choy20194d,schult2023mask3d}. With the maturation of \ac{3dvl} pretraining and instruction tuning techniques, 3D \acp{vlm} have demonstrated significant improvements in capabilities \cite{zhao20213dvg,huang2022multi,abdelreheem20223dreftransformer,chen2022language,huang2022multi,liu2023openshape,xue2024ulip,zhu2025ponderv2,zhu20233d,zhou2024uni3d,wang2025masked,hong20233d,xu2024pointllm,huang2024embodied,yang2024llm,chen2024ll3da,qi2024gpt4point,zhu2024unifying,fu2025scene,zhang2024task,chu2024unified,huang2024chatscene,deng20253d}. In contrast, an alternative line of work leverages 2D perception models to handle \ac{3dvl} tasks and exhibits strong performances \cite{peng2023openscene,shafiullah2023clip,ding2023pla,jatavallabhula2023conceptfusion,huang2023visual,el2024probing,man2024lexicon3d,luo2025dspnet,zhu2024llava,zheng2025video,qi2025gpt4scene}. However, adapting 2D \acp{vlm} to 3D scene understanding poses two challenges: the cost of representing 3D scenes and the absence of effective 3D spatial modeling. Despite recent efforts \cite{zhi2025lscenellm,yu2025inst3d,thai2025splattalk,zheng2025learning,huang2025mllms}, the scene representation still struggles with a performance-efficiency trade-off \cite{huang2025zero,zhang2025adatoken}. In contrast, we propose a novel scene representation equipped with disentangled 3D spatial modeling, which preserves both strong perception capability and high efficiency, delivering an important direction and practical solution for improving the efficiency of 3D VLMs.

\paragraph{3D Vision-Language Datasets and Benchmarks.} Based on indoor 3D scene assets \cite{dai2017scannet,yeshwanth2023scannet++,wald2019rio,mao2022multiscan,baruch2021arkitscenes,chang2017matterport3d,ramakrishnan2021habitat,chang2015shapenet,khanna2024habitat,zheng2020structured3d,fu20213d,gong2023arnold,deitke2022procthor}, existing \ac{3dvl} datasets primarily focus on object grounding \cite{chen2020scanrefer,achlioptas2020referit3d,zhang2023multi3drefer} and \ac{qa} \cite{azuma2022scanqa,ma2023sqa3d} tasks. As the \ac{3dvl} community advances, recent efforts have focused on aggregating diverse scene domains to enable large-scale pretraining \cite{zhu20233d,zhu2024unifying,jia2024sceneverse,wang2024embodiedscan} and instruction tuning \cite{huang2024embodied}, followed by more cross-domain datasets \cite{yang20253d,lyu2024mmscan,linghu2024multi,song2025robospatial,zhang2025flatland} and benchmarks \cite{huang2022perceive,linghu2024multi,huang2025unveiling,yang2025thinking}. Given the potential of cross-domain scaling for \ac{3dvl} learning, we compile a comprehensive \ac{3dvl} data scheme covering four real-world scene domains and five instruction-tuning tasks. We further demonstrate the benefits of domain and task diversity through comprehensive evaluations.

\paragraph{Post-training for Vision-Language Models.} Prior studies \cite{deng2024can,huang2025unveiling,peng2025understanding,zhang2025point} suggest that \ac{sft}, as a common strategy for 3D \acp{vlm}, may undergo weak robustness given the scarce data and overfitting risk in \ac{3dvl} learning. To address this, existing efforts mainly involve data augmentation \cite{yang20253d,kang2024robin3d,zhao2024openscan,linghu2026scenecot}, with the learning objective underexplored. Motivated by the success of \ac{rlhf} \cite{ouyang2022training,schulman2017proximal,rafailov2023direct,yuan2024self,pang2024iterative,liu2024provably,chu2025sft}, recent works \cite{li2024multi,pi2024strengthening,wang2024mdpo,xie2024v} demonstrate the efficacy of \ac{dpo} \cite{rafailov2023direct} on 2D \acp{vlm}. In contrast, we introduce SceneDPO as a novel and effective post-training objective tailored for 3D \acp{vlm}.

\section{Method}\label{sec:method}

In \cref{sec:preliminary}, we first characterize the limitations of current 3D \acp{vlm} by reviewing prior efforts in scene representation, data scaling, and learning objectives. To address these challenges, \cref{sec:model} details our proposed model design, which features a strong yet efficient representation for 3D scenes. Next, \cref{sec:data} outlines our comprehensive \ac{3dvl} instruction-tuning data recipe, spanning four real-world indoor domains and five tasks. Finally, \cref{sec:objective} introduces a novel post-training objective to enhance the robustness of 3D \acp{vlm}.

\subsection{Preliminaries}\label{sec:preliminary}

\paragraph{Capacity-efficiency Trade-off in 3D Scene Representation.} Recent progress in 3D \acp{vlm} has explored several distinct approaches in scene representation, each with inherent trade-offs. Object-centric representations (\eg, 3D-VisTA \cite{zhu20233d}, LEO \cite{huang2024embodied}) process object-centric point clouds using point cloud backbones \cite{qi2017pointnet++} to extract object semantics and spatial modules \cite{chen2022language} to model inter-object spatial relations. However, they rely heavily on accurate object masks and fall short in capturing global scene context. In contrast, query-based representations (\eg, 3D-LLM \cite{hong20233d}, PQ3D \cite{zhu2024unifying}) employ learnable queries to extract scene features, but can struggle with optimization stability and may overlook fine-grained spatial details. More recently, video-based representations (\eg, Video-3D LLM \cite{zheng2025video}, GPT4Scene \cite{qi2025gpt4scene}) take multi-view images or videos as input, relying on 3D position encoding or temporal information for 3D awareness. While video-based representations prevail due to the strong perception capacity of 2D \acp{vlm}, they incur substantial token overhead that severely limits training efficiency and data scalability. Therefore, developing a strong yet efficient representation is a critical necessity for current 3D \acp{vlm}.

\paragraph{Fragmented 3D-VL Data Across Scenes and Tasks.} In addition to architectural refinements, significant efforts have focused on scaling \ac{3dvl} data across diverse scene domains and tasks. For instance, LEO \cite{huang2024embodied} designs five \ac{3dvl} tasks (\eg, captioning, reasoning, planning, and dialogue) across two real-world scene domains, ScanNet \cite{dai2017scannet} and 3RScan \cite{wald2019rio}. In contrast, SceneVerse \cite{jia2024sceneverse} scales \ac{3dvl} data extensively across more diverse scene domains such as MultiScan \cite{mao2022multiscan} and HM3D \cite{ramakrishnan2021habitat} while primarily focusing on object grounding tasks. Similarly, MSR3D \cite{linghu2024multi} advances the situated reasoning task across multiple 3D scene domains. Nonetheless, these efforts have not delivered a unified comprehensive data recipe for the development of more general and robust 3D \acp{vlm}.

\paragraph{Insufficient Learning Objective for Robust 3D \acp{vlm}.} The prevailing paradigm for training 3D \acp{vlm} relies on auto-regressive language modeling, where an \ac{llm} generates textual responses through next-token prediction. The learning objective is usually formulated as \ac{sft}, \ie, imitating the ground-truth labels. However, recent breakthroughs in \acp{llm} and \acp{vlm} suggest that relying on \ac{sft} alone can be insufficient for building robust models, even if equipped with comprehensive \ac{sft} data \cite{ouyang2022training,jaech2024openai-o1,guo2025deepseekr1,team2025kimik1.5,linghu20263drft}. In particular, the scarcity and complexity of \ac{3dvl} data can enhance the overfitting risks and thereby highlight the demand of effective post-training objective for 3D \acp{vlm}. 

\subsection{Model}
\label{sec:model}

\begin{figure*}[t!]
    \centering
    \includegraphics[width=\linewidth]{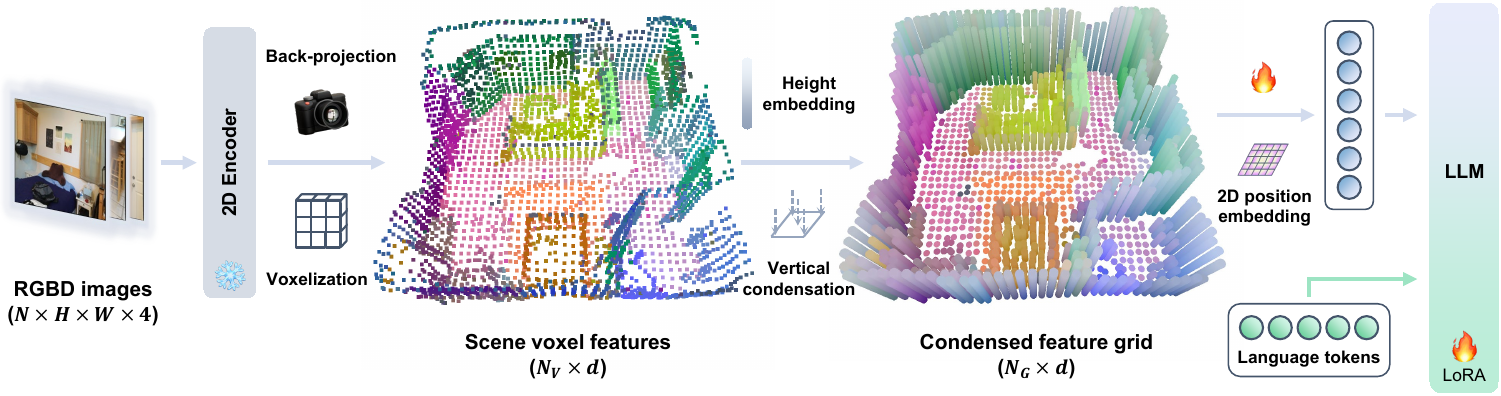}
    \caption{\textbf{\model model design.} \model extracts 2D visual features from multi-view RGB-D frames and transforms the features into a \acf{cfg}, significantly reducing the token overhead while preserving 3D spatial structure. An \ac{llm} performs auto-regressive language modeling based on the \ac{cfg} tokens and language tokens.}
    \label{fig:model}
\end{figure*}

\paragraph{Overview.} As illustrated in \cref{fig:model}, given multi-view RGB-D images, \model extracts 2D visual features and transforms them into a \acf{cfg}. An LLM then performs auto-regressive language modeling over the joint sequence of \ac{cfg} tokens and language tokens. Compared to previous object-centric 3D \acp{vlm} \cite{huang2024embodied,zhu2024unifying,huang2024chatscene,linghu2024multi}, \model leverages 2D perception to address limitations such as the dependency of object masks and insufficient perception capacity. In contrast to recent video-based \acp{vlm} \cite{zhu2024llava,zheng2025video,qi2025gpt4scene}, \model substantially reduces the representation overhead by abstracting the 3D scene into a structured grid, simplifying 3D spatial modeling while improving efficiency.

\paragraph{2D Perception and Back-projection.} We employ a 2D visual encoder to process $N$ multi-view RGB images $\{I_n \in \mathbb{R}^{H\times W\times 3} \}_{n=1}^N$, yielding feature maps $\{M_n \in \mathbb{R}^{h\times w\times d}\}_{n=1}^N$, where $d$ denotes the feature hidden dimension. Corresponding depth maps $\{D_n \in \mathbb{R}^{H\times W}\}_{n=1}^N$ are downsampled to match the feature resolution $h\times w$. Each feature patch at coordinate $u = (i,j)$ is back-projected into a 3D point based on the depth value and camera parameters. Let $K \in \mathbb{R}^{3\times 3}$ and $T \in \mathbb{R}^{4\times 4}$ denote the intrinsic and extrinsic camera matrices. Let $p = (x,y,z)$ denote the 3D world coordinate, the homogeneous 3D coordinate $\bar{p} = [x,y,z,1]^\top$ is derived from the homogeneous 2D coordinate $\bar{u} = [i,j,1]^\top$ as:
\begin{equation}
\bar{p} = \begin{bmatrix} p \\ 1
\end{bmatrix} = T \begin{bmatrix} D(u)K^{-1}\bar{u} \\ 1 \end{bmatrix}.
\end{equation}
The resulting 3D points $\mathcal{P}$ inherit 2D features from $\{M_n\}$, where each point $p \in \mathcal{P}$ is associated with a feature $f_p \in \mathbb{R}^d$.

\paragraph{Voxelization.} We first voxelize the 3D points by averaging point features within each voxel, with empty voxels discarded. For referring tasks such as object captioning, a learnable anchor embedding is added to voxels that fall within the referred region. This yields an initial voxel-based representation $v_{(x,y,z)}$, which remains computationally expensive for representation and challenging for 3D spatial modeling.

\paragraph{Condensed Feature Grid.} To further improve efficiency, our core is to condense voxels into a 2D planar grid by pooling voxel features vertically within each pillar. To address the challenge of 3D spatial modeling, we disentangle the position encoding along vertical and horizontal directions, respectively. The process from voxels to \ac{cfg} consists of three stages:
\begin{itemize}[leftmargin=*]
    \item \textbf{Vertical Position Encoding.} We encode the vertical position (\ie, height) of each voxel using \ac{rope} \cite{su2024roformer}, which applies a rotation-like transformation to the feature. The rotation angles relate to a pre-defined channel-wise frequencies and position value (\ie, height). Compared to additive position embeddings \cite{vaswani2017attention,tancik2020fourier}, the rotary nature enables \ac{rope} to better capture vertical spatial relations, such as distinguishing objects at different heights (see example in \supp). The height-encoded voxel feature $\bar{v}_{(x,y,z)}$ is formulated as:
    \begin{equation}
    \bar{v}_{(x,y,z)} = R(z)v_{(x,y,z)}.
    \end{equation}

    \item \textbf{Vertical Condensation.} Let $C(x^*,y^*)$ denote the count of voxels located in the pillar at $(x^*,y^*)$. For each horizontal position $(x^*,y^*)$ where $C(x^*,y^*) > 0$, the conversion from voxels $\bar{v}_{(x,y,z)}$ to \ac{cfg} token $g_{(x^*,y^*)}$ is formulated as:
    \begin{equation}
    g_{(x^*,y^*)} = \frac{1}{C(x^*,y^*)} \sum_{(x,y,z)} \bar{v}_{(x,y,z)} \mathds{1}(x=x^*,y=y^*).
    \end{equation}

    \item \textbf{Horizontal Position Encoding.} We encode the horizontal position $(x^*,y^*)$ with 2D Fourier features \cite{tancik2020fourier}, which applies a linear transformation (denoted by $W$) to the position value $(x^*,y^*)$ and then computes sinusoidal components, followed by an MLP to enhance expressiveness. Let $\mathbf{p}=[x^*,y^*]^\top$ denote the horizontal position, the final \ac{cfg} token $\bar{g}_{(x^*,y^*)}$ is formulated as:
\end{itemize}
\small{
\begin{equation}
\bar{g}_{(x^*,y^*)} = g_{(x^*,y^*)} + \text{MLP}\big(\left[\cos{(2\pi \mathbf{p}^\top W)} \parallel \sin{(2\pi \mathbf{p}^\top W})\right]^\top \big).
\end{equation}
}

\paragraph{Large Language Model.} An \ac{llm} auto-regressively generates text output based on the joint sequence of \ac{cfg} tokens $\bar{g}_{(x^*,y^*)}$ and instruction tokens. The \ac{cfg} scene tokens are directly used without any resampling process \cite{zhu2024unifying,chen2024ll3da} owing to low token overhead.

\subsection{Data}
\label{sec:data}

The construction of \model training data is guided by three principles: (1) the diversity of scene domains, which demonstrates crucial for \ac{3dvl} learning \cite{jia2024sceneverse}; (2) the diversity of tasks, which we think necessary given the benefits of long-response tasks \cite{huang2024embodied,lyu2024mmscan}; and (3) the balance between quality and scale, considering the potential risks in hindering model performance from low-quality data. We detail the composition of our data in \cref{tab:data_components}.

\paragraph{Scene Domains.} We include four real-world indoor domains: ScanNet \cite{dai2017scannet}, 3RScan \cite{wald2019rio}, MultiScan \cite{mao2022multiscan}, and ARKitScenes \cite{baruch2021arkitscenes}, following prior practices in scaling 3D scenes \cite{jia2024sceneverse,linghu2024multi}. We exclude scene domains that lack attribute-rich scene graphs (\eg, HM3D \cite{ramakrishnan2021habitat}), which are essential for LLM-assisted data generation.

\begin{table*}[t!]
\centering
\caption{\textbf{Overview of \model training data}. The entries include: ``SV'' for SceneVerse \cite{jia2024sceneverse}, ``MM'' for MMScan \cite{lyu2024mmscan}, ScanQA \cite{azuma2022scanqa}, ``SQA'' for SQA3D \cite{ma2023sqa3d}, MSQA \cite{linghu2024multi}, LEO \cite{huang2024embodied}, ``\cmark'' for newly created data in this work, and ``-'' for filtered data due to quality control.}
\resizebox{\linewidth}{!}{
\begin{tabular}{lccccccc}
\toprule
      & ScanNet & 3RScan & MultiScan & ARKitScenes & Overall count & Avg. length (str) & Avg. length (words) \\ 
     \midrule
     ObjCap & SV, MM & LEO, SV, MM & SV & SV & 216k & 299 & 56 \\
     SceneCap & MM, \cmark & LEO, MM, \cmark & \cmark & \cmark & 128k & 633 & 104 \\
     QA & ScanQA, SQA, MSQA & MSQA & - & MSQA & 289k & 31 & 6 \\
     Plan & \cmark & LEO & \cmark & \cmark & 18k & 534 & 96 \\
     Dialog & \cmark & LEO & \cmark & \cmark & 61k & 93 & 18 \\
     \bottomrule
\end{tabular}
}
\label{tab:data_components}
\end{table*}

\paragraph{Tasks and Datasets.} We incorporate five prevalent \ac{3dvl} tasks with natural language outputs, which are compatible with a unified instruction tuning framework. We exclude the 3D object grounding task due to its different formulation. We think the absence of the grounding task is not detrimental, given the success of 2D \acp{vlm} without explicit grounding supervision \cite{liu2023visual,openai2023gpt4,wang2024qwen2,chen2024internvl}. In addition to existing datasets, we generate new data by prompting LLMs with scene graphs \cite{huang2024embodied,linghu2024multi} to fill the blank (``\cmark'' entries) in \cref{tab:data_components}. Our data covers five categories of text-output tasks:
\begin{itemize}[nolistsep,leftmargin=*]
    \item \textbf{Object Captioning.} This task is to describe a specific object in natural language. We adopt object captions from SceneVerse \cite{jia2024sceneverse} and MMScan \cite{lyu2024mmscan}. We exclude datasets that re-purpose object grounding texts as captions \cite{chen2021scan2cap,achlioptas2020referit3d}, as they lack diverse descriptions regarding object attributes.
    \item \textbf{Scene Captioning.} This task requires generating comprehensive descriptions of 3D scenes. We use scene captions from LEO \cite{huang2024embodied} and MMScan \cite{lyu2024mmscan}, and generate situated scene captions that incorporate situations to resolve spatial ambiguities (\eg, left and right).
    \item \textbf{Question Answering.} This task requires answering general questions about the scene. We include ScanQA \cite{azuma2022scanqa}, SQA3D \cite{ma2023sqa3d}, and MSQA \cite{linghu2024multi}. We exclude newly-generated QA data as we find it can exhibit trivial patterns and degrade model performance.
    \item \textbf{Planning.} This task is to generate a step-by-step grounded plan for a high-level goal (\eg, ``organize the room''). We use the planning data from LEO \cite{huang2024embodied} for 3RScan, and generate new data for other scene domains.
    \item \textbf{Dialogue.} This task concerns generating responses conditioned on both the 3D scene and dialogue context. We use the dialogue data from LEO \cite{huang2024embodied} for 3RScan, and generate new data for other scene domains.
\end{itemize}

\subsection{Post-training}
\label{sec:objective}

We argue that only \ac{sft} can be insufficient to build robust 3D \acp{vlm}, as evidenced by prior studies \cite{li2024multi,pi2024strengthening,wang2024mdpo}. In particular, given the pronounced risk of overfitting in 3D \acp{vlm} \cite{deng2024can,huang2025unveiling}, it is crucial to design an effective post-training objective to enhance their robustness.

To this end, we propose SceneDPO, a novel post-training objective for 3D \acp{vlm}. We start with a DPO-like loss term $\Ls_a$ that contrasts between positive answer $a_\text{\cmark}$ and negative answer $a_\text{\xmark}$. Motivated by the issue of visual ignorance \cite{huang2025unveiling}, we introduce a loss term $\Ls_s$ that contrasts between positive scene $s_\text{\cmark}$ and negative scene $s_\text{\xmark}$. This discourages the model from predicting the current answer when conditioned on irrelevant scenes. To mitigate degradation of positive answers, we incorporate a negative log-likelihood loss term $\Ls_\text{NLL}$, which proves critical in prior works \cite{liu2024provably,pang2024iterative,wang2024mdpo}. Given a training tuple $(s_\text{\cmark},s_\text{\xmark},q,a_\text{\cmark},a_\text{\xmark})\sim \mathcal{D}$, the overall loss $\Ls$ is defined as follows:
\begin{equation}
\Ls_a = - \mathbb{E}_{\mathcal{D}} \log \sigma \bigg( \beta_a \big[ r(a_\text{\cmark}, s_\text{\cmark}, q) - r(a_\text{\xmark}, s_\text{\cmark}, q) \big] \bigg),
\end{equation}
\begin{equation}
\Ls_s = - \mathbb{E}_{\mathcal{D}} \log \sigma \bigg( \beta_s \big[ r(a_\text{\cmark}, s_\text{\cmark}, q) - r(a_\text{\cmark}, s_\text{\xmark}, q) \big] \bigg),
\end{equation}
\begin{equation}
\Ls_{\text{NLL}} = - \mathbb{E}_{\mathcal{D}} \log \pi_\theta (a_{\text{\cmark}} \mid s_{\text{\cmark}}, q),
\end{equation}
\begin{equation}
\mathcal{L} = w_a \mathcal{L}_a + w_s \mathcal{L}_s + \mathcal{L}_{\text{NLL}},
\end{equation}
where $r(a, s, q) = \log \frac{\pi_\theta(a \mid s, q)}{\pi_{\text{ref}}(a \mid s, q)}$ denotes the log-ratio between the policy ($\theta$) and reference model (\text{ref}), $\sigma$ denotes the sigmoid function, and $w_a$, $w_s$, $\beta_a$, and $\beta_s$ are scalar hyperparameters.

\section{Experiment}
\label{sec:exp}

In \cref{sec:results_qa,sec:results_cap_grd}, we first compare \model against \sota models to highlight its performance and efficiency, covering various \ac{3dvl} tasks and benchmarks. In \cref{sec:model_ablation}, we present ablation studies on model design to show the effectiveness of \ac{cfg}. In \cref{sec:data_ablation}, we analyze various training data configurations to reveal the influence of task diversity, scene domain coverage, data quality, and data scale on \ac{3dvl} learning. In \cref{sec:post_training}, we further explore the effect of SceneDPO and compare it to other post-training alternatives.

\paragraph{Implementation Details.} We initialize the 2D visual encoder and LLM with Qwen2.5-VL-7B \cite{bai2025qwen2}. The learnable parameters include position embedding, anchor embedding, and LoRA parameters \cite{hu2022lora} of the LLM, which amount to 66M in total. We set the voxel size to 0.2 meters and retain up to 750 scene tokens for \ac{cfg}. We train \model on our comprehensive \ac{3dvl} dataset for 5 epochs, which takes 2 days with 8 NVIDIA A100 80G GPUs. We adopt AdamW optimizer \cite{loshchilov2017decoupled} with a base learning rate at $3\times 10^{-5}$, scheduled with linear warmup and cosine decay. We provide detailed hyperparameters in \supp.

\paragraph{Evaluation.} Our evaluation primarily focuses on 3D QA task, including ScanQA \cite{azuma2022scanqa} for general QA, SQA3D \cite{ma2023sqa3d} for situated QA, MSQA \cite{linghu2024multi} for situated QA across multiple scene domains, and Beacon3D \cite{huang2025unveiling} for robust QA evaluation in diverse scene domains. To further demonstrate \model's versatility, we report finetuning results on Scan2Cap \cite{chen2021scan2cap} for 3D object captioning and ScanRefer \cite{chen2020scanrefer} for 3D object grounding. Following established conventions, we report n-gram metrics for ScanQA and Scan2Cap, exact-match (EM) accuracy for SQA3D, detailed GPT-Score for MSQA and Beacon3D, and accuracy at \ac{iou} thresholds of $0.25$ (Acc@$0.25$) and $0.5$ (Acc@$0.5$) for ScanRefer. We denote \model before post-training as our default configuration. A detailed analysis of post-training is provided separately in \cref{sec:post_training}. For cross-domain benchmarks, we use different colors to denote the scene domain: \textcolor{scannet}{ScanNet}, \textcolor{3rscan}{3RScan}, \textcolor{multiscan}{MultiScan}, and \textcolor{arkitscenes}{ARKitScenes}. The main results are reported on \textcolor{scannet}{ScanNet} as it is the only domain with sufficient baselines for comparison. We provide evaluations on other domains in \cref{sec:data_ablation,sec:post_training} for cross-domain analysis. 

\subsection{Main Results on Question Answering}
\label{sec:results_qa}

\paragraph{Baselines.} For comparison, we include \sota 3D \acp{vlm} across four categories:
(1) query-based methods, ranging from early work like ScanQA \cite{azuma2022scanqa} to recent 3D-LLaVA \cite{deng20253d}; (2) object-centric methods, from early work like 3D-VisTA \cite{zhu20233d} to recent Inst3D-LMM \cite{yu2025inst3d}; (3) voxel-based methods, such as Scene-LLM \cite{fu2025scene} and LLaVA-3D \cite{zhu2024llava}; and (4) video-based methods, including Video-3D LLM \cite{zheng2025video} and GPT4Scene \cite{qi2025gpt4scene}. We also show their scene representations and associated costs. For Scene-LLM, which only reports the voxel resolution (0.18m), we estimate the token count based on LLaVA-3D (3096 tokens at 0.2m resolution) as follows: $0.2^3 \times 3096 \div 0.18^3 \simeq 4247$. For video-based methods, we derive the token count assuming a standard input of 32 frames.

\paragraph{Main Results.} As shown in \cref{tab:overall_results,tab:msqa,tab:beacon3d}, \model achieves \sota performance on most 3D QA benchmarks with a much lower representation cost. Specifically, \model shows superior performance in situated reasoning tasks, including SQA3D and MSQA. \model also exhibits notable strength in spatial-reasoning tasks, including spatial and navigation categories in MSQA and the spatial category in Beacon3D. While video-based models achieve competitive performance, their representations require numerous scene tokens and limit efficiency, hindering scalable \ac{3dvl} learning across a broader range of scene domains. In contrast, the efficient representation of \model enables cross-domain scaling of \ac{3dvl} data, which yields consistent performance improvements (\cref{sec:data_ablation}). These results underscore the advantages of our representation in perception capability, efficiency, and data scalability.

\begin{table}[t!]
\centering
\captionof{table}{\textbf{Scene representations and results on ScanQA and SQA3D.} ``C'' stands for CIDEr, ``B-4'' for BLEU-4, ``EM'' for exact-match accuracy, and ``EM-R'' for refined exact-match accuracy \cite{huang2024embodied}. \textcolor{gray}{Number} indicates scene token count estimated by voxel resolution. Benchmarks are colorized according to scene domain (\textcolor{scannet}{ScanNet}).}
\vspace{0.2em}
\resizebox{\linewidth}{!}{
\begin{tabular}{llcccc}
    \toprule
     \multirow{2}{*}{\raisebox{-0.8ex}{Model}} & \multirow{2}{*}{\raisebox{-0.8ex}{Scene (\#token)}} & \multicolumn{2}{c}{\textcolor{scannet}{ScanQA (val)}} & \multicolumn{2}{c}{\textcolor{scannet}{SQA3D (test)}} \\
     \cmidrule(lr){3-4} \cmidrule(lr){5-6}
     & & C & B-4 & EM & EM-R \\
    \midrule
    ScanQA \cite{azuma2022scanqa} & Query (256) & 64.9 & 10.1 & 47.2 & - \\
    3D-LLM \cite{hong20233d} & Query (32) & 74.5 & 12.9 & 49.8 & - \\
    PQ3D \cite{zhu2024unifying} & Query (80) & - & - & 47.1 & - \\
    DSPNet \cite{luo2025dspnet} & Query (256) & - & - & 50.4 & - \\
    3D-LLaVA \cite{deng20253d} & Query (100) & 92.6 & \textbf{17.1} & 54.5 & 56.6 \\
    \midrule
    3D-VisTA \cite{zhu20233d} & Object (80) & 69.6 & 10.4 & 48.5 & - \\
    LEO \cite{huang2024embodied} & Object (60) & \textbf{101.4} & 13.2 & 50.0 & 52.4 \\
    SceneVerse \cite{jia2024sceneverse} & Object (80) & - & - & 49.9 & - \\
    Chat-Scene \cite{huang2024chatscene} & Object (200) & 87.7 & 14.3 & 54.6 & 57.5 \\
    Inst3D-LMM \cite{yu2025inst3d} & Object (200) & 88.6 & 14.9 & - & - \\
    \midrule
    Scene-LLM \cite{fu2025scene} & Voxel (\textcolor{gray}{4247}) & 80.0 & 11.7 & 53.6 & - \\
    LLaVA-3D \cite{zhu2024llava} & Voxel (3096) & 91.7 & 14.5 & 55.6 & 57.6 \\
    \midrule
    Video-3D LLM \cite{zheng2025video} & Video (6720) & 100.5 & 16.3 & 57.7 & - \\
    GPT4Scene \cite{qi2025gpt4scene} & Video (8262) & 96.3 & 15.5 & 59.4 & 62.4 \\
    \midrule

    \model & Grid (750) & 100.4 & 15.5 & \textbf{60.8} & \textbf{63.7} \\

    \bottomrule
\end{tabular}
}
\label{tab:overall_results}
\end{table}

\begin{table}[t!]
\centering
\captionof{table}{\textbf{Detailed results on MSQA (\textcolor{scannet}{ScanNet}) test set.} $^*$ indicates text-only input. ``Count.'' stands for counting, ``Exist.'' for existence, ``Attr.'' for attribute, and ``Navi.'' for navigation. Performances are evaluated under GPT-Score metrics.}
\vspace{0.2em}
\resizebox{\linewidth}{!}{
\begin{tabular}{lcccccccc}
    \toprule
     Model & Count. & Exist. & Attr. & Spatial & Navi. & Others & Overall \\
    \midrule
    GPT-4o$^*$ \cite{openai2023gpt4} & 32.3 & 79.3 & \textbf{79.0} & 37.0 & 31.7 & \textbf{91.6} & 52.3 \\
    LEO \cite{huang2024embodied} & 32.5 & 88.5 & 58.7 & 44.2 & 39.6 & 81.4 & 54.8 \\
    MSR3D \cite{linghu2024multi} & 32.3 & \textbf{93.1} & 50.0 & 46.5 & 54.1 & 75.6 & 54.2 \\
    SplatTalk \cite{thai2025splattalk} & 19.6 & 60.3 & 44.0 & 35.8 & 35.5 & 61.8 & 41.8 \\
    \midrule
    \model & \textbf{39.3} & 92.7 & 56.9 & \textbf{59.3} & \textbf{59.7} & 82.8 & \textbf{61.7} \\
    \bottomrule
\end{tabular}
}
\label{tab:msqa}
\end{table}

\begin{table}[t!]
\centering
\captionof{table}{\textbf{Detailed results on Beacon3D (\textcolor{scannet}{ScanNet}).} $^*$ indicates text-only input. ``App.'' stands for appearance, ``Geo.'' for geometry, ``Spa.'' for spatial, ``Exi.'' for existence, and ``Obj.'' for object-centric metrics \cite{huang2025unveiling}. Performances are evaluated under GPT-Score metrics.}
\vspace{0.2em}
\resizebox{\linewidth}{!}{
\begin{tabular}{lcccccccc}
    \toprule
    \multirow{2}{*}{\raisebox{-0.8ex}{Model}} & \multirow{2}{*}{\raisebox{-0.8ex}{Class}} & \multirow{2}{*}{\raisebox{-0.8ex}{App.}} & \multirow{2}{*}{\raisebox{-0.8ex}{Geo.}} & \multirow{2}{*}{\raisebox{-0.8ex}{Spa.}} & \multirow{2}{*}{\raisebox{-0.8ex}{Exi.}} & \multicolumn{2}{c}{Overall} \\
    \cmidrule(lr){7-8}
    & & & & & & Case & Obj. \\
    \midrule
    3D-VisTA \cite{zhu20233d} & 28.4 & 35.7 & 41.6 & 48.0 & 55.0 & 43.2 & 7.3 \\
    PQ3D \cite{zhu2024unifying} & 37.8 & 45.8 & 32.1 & 19.2 & 44.5 & 35.9 & 4.2 \\
    SceneVerse \cite{jia2024sceneverse} & 26.4 & 40.4 & 40.0 & 35.0 & 54.1 & 40.5 & 4.7 \\
    LEO \cite{huang2024embodied} & 16.4 & 39.8 & 47.6 & 52.8 & 54.3 & 45.2 & 7.5 \\
    Chat-Scene \cite{huang2024chatscene} & 30.0 & 42.7 & 50.0 & 53.9 & 62.9 & 49.8 & 10.9 \\
    GPT-4o$^*$ \cite{openai2023gpt4} & 39.2 & 49.9 & 53.8 & 58.4 & \textbf{70.0} & 56.0 & 15.3 \\
    LLaVA-3D \cite{zhu2024llava} & 35.1 & 66.7 & \textbf{62.5} & 54.2 & 62.9 & 59.1 & 19.0 \\
    Video-3D LLM \cite{zheng2025video} & 40.1 & 64.1 & 60.6 & 55.3 & 64.1 & 59.0 & 17.9 \\
    GPT4Scene \cite{qi2025gpt4scene} & 38.1 & 59.7 & 59.3 & 52.6 & 66.1 & 57.2 & 17.9 \\
    \midrule
    \model & \textbf{41.2} & \textbf{67.4} & 57.0 & \textbf{61.0} & 56.7 & \textbf{59.5} & \textbf{19.2} \\
    \bottomrule
\end{tabular}
}
\label{tab:beacon3d}
\end{table}

\subsection{Finetuning Results on Captioning and Grounding}
\label{sec:results_cap_grd}

\paragraph{Object Captioning.} \model inherently supports object captioning by incorporating the anchor embedding, which resembles a click prompt at the target location and conditions the model to generate descriptions for the target object. We finetune \model on Scan2Cap and report the evaluation results in \cref{tab:scan2cap}. \model achieves the highest scores across two of the four n-gram metrics and the highest average score, establishing \sota performance on Scan2Cap and demonstrating strong capability in object captioning.

\begin{table}[t!]
\centering
\captionof{table}{\textbf{Finetuning results on Scan2Cap val set.} ``C'' stands for CIDEr, ``B-4'' for BLEU-4, ``M'' for METEOR, and ``R'' for ROUGE. For methods that rely on object proposals (\eg, Mask3D \cite{schult2023mask3d}), we use the IoU@0.5 criterion for comparison.}
\vspace{0.2em}
\resizebox{\linewidth}{!}{
\begin{tabular}{lccccc}
    \toprule
     Model & C & B-4 & M & R & Average \\
    \midrule
    Scan2Cap \cite{chen2021scan2cap} & 35.2 & 22.4 & 21.4 & 43.5 & 30.6 \\
    3DJCG \cite{cai20223djcg} & 49.5 & 31.0 & 24.2 & 50.8 & 38.9 \\
    3D-VisTA \cite{zhu20233d} & 66.9 & 34.0 & 27.1 & 54.3 & 45.6 \\
    Vote2Cap-DETR++ \cite{chen2024vote2cap} & 74.4 & 37.2 & 26.2 & 53.3 & 47.8 \\
    3DVLP \cite{zhang2024vision} & 54.4 & 34.1 & \textbf{34.3} & 54.3 & 44.3 \\
    LEO \cite{huang2024embodied} & 72.4 & 38.2 & 27.9 & 58.1 & 49.2 \\
    LL3DA \cite{chen2024ll3da} & 65.2 & 36.8 & 26.0 & 55.1 & 45.8 \\
    PQ3D \cite{zhu2024unifying} & 80.3 & 36.0 & 29.1 & 57.9 & 50.8 \\
    Chat-Scene \cite{huang2024chatscene} & 77.2 & 36.3 & 28.0 & 58.1 & 49.9 \\
    LLaVA-3D \cite{zhu2024llava} & 79.2 & 41.1 & 30.2 & 63.4 & 53.5 \\
    3D-LLaVA \cite{deng20253d} & 78.8 & 36.9 & 27.1 & 57.7 & 50.1 \\
    Video-3D LLM \cite{zheng2025video} & 80.0 & 40.2 & 28.5 & 61.7 & 52.6 \\
    GPT4Scene \cite{qi2025gpt4scene} & \textbf{86.3} & 40.6 & 28.2 & 59.3 & 53.6 \\
    \midrule
    \model & 77.3 & \textbf{43.8} & 30.3 & \textbf{64.9} & \textbf{54.1} \\
    \bottomrule
\end{tabular}
}
\label{tab:scan2cap}
\end{table}

\paragraph{Object Grounding.} \model can be naturally adapted for object grounding by directly generating 3D bounding boxes in a structured text format $[x_{\text{center}}, y_{\text{center}}, z_{\text{center}}, l_x, l_y, l_z]$, utilizing the anchor embedding similar to prior works \cite{chen2024ll3da,zhu2024llava}. When object proposals are available, we can further refine the prediction by retrieving the proposal with the maximum \ac{iou} with the predicted bounding box. We finetune \model on ScanRefer and report performance both with and without Mask3D proposals \cite{schult2023mask3d} in \cref{tab:grounding}. The results demonstrate \model's competitive performance despite the inherent difficulty of directly predicting 3D bounding boxes. Specifically, the strong Acc@0.25 confirms the effectiveness of our 3D spatial encoding, while the relatively weaker Acc@0.5 suggests that representation compression may limit the precision of fine-grained localization. Notably, integrating object proposals, which simplifies the grounding task into a selection task as seen in prior works \cite{huang2024chatscene,zheng2025video}, greatly refines the final prediction and significantly improves the \sota accuracies.

\begin{table}[t!]
\centering
\captionof{table}{\textbf{Finetuning results on ScanRefer val set.} We compare to previous LLM-based grounding methods, reporting results both with and without the use of Mask3D object proposals \cite{schult2023mask3d}.}
\vspace{0.2em}
\resizebox{\linewidth}{!}{
\begin{tabular}{lcc}
    \toprule
    Model & Acc@0.25 & Acc@0.5 \\
    \midrule
    ReGround3D (w/o Mask3D) \cite{zhu2024scanreason} & 53.1 & 41.1 \\
    LLaVA-3D (w/o Mask3D) \cite{zhu2024llava} & 50.1 & 42.7 \\
    Chat-Scene (w/ Mask3D) \cite{huang2024chatscene} & 55.5 & 50.2 \\
    Video-3D LLM (w/ Mask3D) \cite{zheng2025video} & 57.9 & 51.2 \\
    \midrule
    LEO-VL (w/o Mask3D) & 60.0 & 20.6 \\
    LEO-VL (w/ Mask3D) & 83.7 & 75.2 \\
    \bottomrule
\end{tabular}
}
\label{tab:grounding}
\end{table}

\subsection{Model Analysis}
\label{sec:model_ablation}

\begin{figure*}[t!]
\centering
\begin{minipage}{0.52\linewidth}
    \centering
    \includegraphics[width=\linewidth]{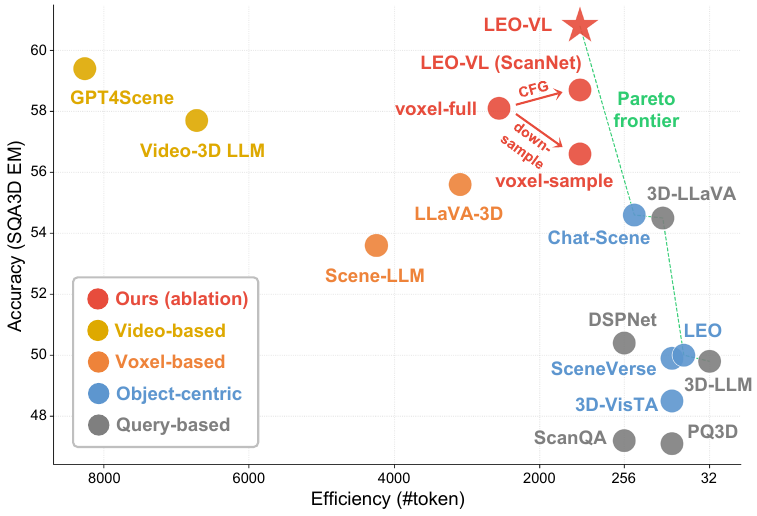}
    \captionof{figure}{\textbf{Joint visualization of accuracy and efficiency.} Accuracy is measured by exact-match (EM) accuracy on SQA3D, while efficiency is measured by scene token count. \model reaches a Pareto optimum between efficiency and accuracy.}
    \label{fig:voxel_ablation}
\end{minipage}
\hfill
\begin{minipage}{0.46\linewidth} 
    \centering
    \vspace{0.5em}
    \includegraphics[width=\linewidth]{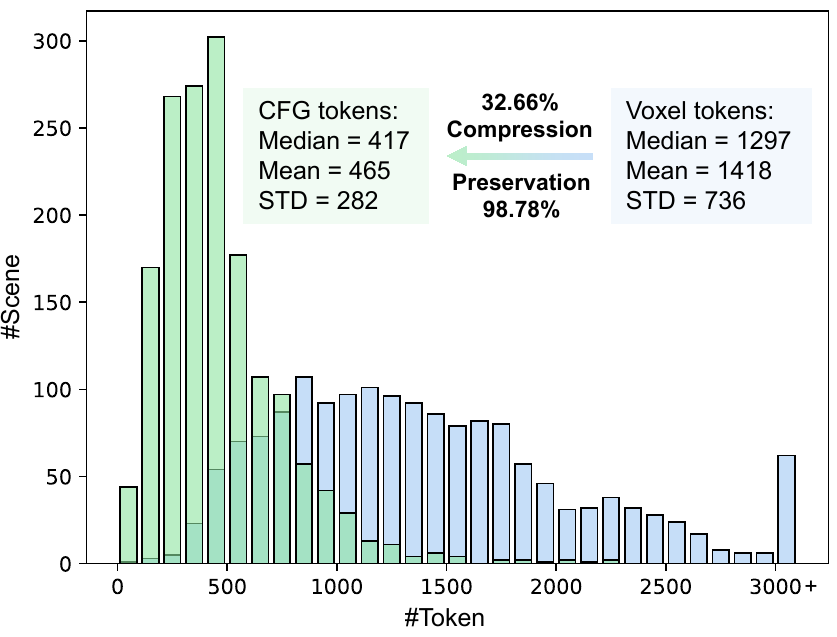}
    \captionof{figure}{\textbf{Statistics of scene token count on ScanNet.} Blue bars denote voxel tokens and green bars denote \ac{cfg} tokens after vertical condensation.}
    \label{fig:voxel_stats}
\end{minipage}
\end{figure*}

\paragraph{Accuracy \vs Efficiency.} We provide a joint visualization of accuracy and efficiency in \cref{fig:voxel_ablation}. Accuracy is measured using the EM on SQA3D, the metric with the most available reference results, while efficiency is measured by the token count of scene representation. As shown in \cref{fig:voxel_ablation}, \model achieves a Pareto optimum between accuracy and efficiency. To further demonstrate the efficiency of \ac{cfg}, we conduct an ablation study by training on the ScanNet subset, comparing \model (ScanNet) with two representation alternatives: \textbf{voxel-full}, which retains all voxel tokens without condensation; and \textbf{voxel-sample}, which simply downsamples voxel tokens to match the token count of \ac{cfg}. The results in \cref{fig:voxel_ablation} show that compared to voxel-full, voxel-sample suffers a drop in accuracy, while \ac{cfg} even slightly improves the accuracy. These results suggest that high accuracy in \ac{3dvl} tasks can be achieved with significantly lower token costs, as enabled by our efficient \ac{cfg} representation.

\paragraph{Statistics of Scene Token Count.} In \cref{fig:voxel_stats}, we present statistics of the number of voxel tokens and \ac{cfg} tokens on ScanNet, with their difference showing the effect of vertical condensation. In addition, we report two metrics: \textbf{compression rate}, defined as the ratio of \ac{cfg} token count to the raw voxel token count before condensation; and \textbf{preservation rate}, defined as the proportion of voxels retained by the final \ac{cfg} tokens (clipped to 750). The results show that \ac{cfg} achieves an average compression rate of about 33\% while preserving nearly 99\% of the scene tokens, which demonstrates the high efficiency of our \ac{cfg} representation.

\paragraph{Backbone Capability.} To isolate the advantages of \ac{cfg} representation from backbone capability, we conduct an ablation study on the ScanNet subset. We compare \model (ScanNet) against two baselines trained on identical data: LEO \cite{huang2024embodied} and a vanilla Qwen2.5-VL-7B \cite{bai2025qwen2}, \ie, the backbone of \model without \ac{cfg}. As shown in \cref{tab:backbone_ablation}, \model (ScanNet) consistently outperforms both baselines across all benchmarks. In particular, \model exhibits both significantly higher performance and efficiency compared to the vanilla Qwen2.5-VL-7B. These results underscore the improvements from our architectural design rather than backbone capability or data scaling.

\begin{table*}[t!]
\centering
\begin{minipage}{0.66\linewidth}
\centering
\captionof{table}{\textbf{Results of backbone ablation on the ScanNet subset.} Baselines are trained on identical data (ScanNet subset). $^\text{\textdagger}$ indicates the backbone of \model without \ac{cfg}. Benchmark metrics follow \cref{tab:overall_results,tab:msqa,tab:beacon3d}.}
\resizebox{\linewidth}{!}{
\begin{tabular}{lccccccc}
    \toprule
    \multirow{2}{*}{\raisebox{-0.8ex}{Model}} & \multicolumn{2}{c}{\textcolor{scannet}{ScanQA}} & \multicolumn{2}{c}{\textcolor{scannet}{SQA3D}} & \textcolor{scannet}{MSQA} & \multicolumn{2}{c}{\textcolor{scannet}{Beacon3D}} \\
     \cmidrule(lr){2-3} \cmidrule(lr){4-5} \cmidrule(lr){6-6} \cmidrule(lr){7-8}
     & C & B-4 & EM & EM-R & Overall & Case & Obj. \\
    \midrule
    LEO \cite{huang2024embodied} & 80.3 & 12.2 & 49.4 & 51.8 & 53.4 & 48.4 & 9.4 \\
    Qwen2.5-VL-7B$^\text{\textdagger}$ \cite{bai2025qwen2} & 96.6 & 16.0 & 54.1 & 57.1 & 56.1 & 55.4 & 17.4 \\
    \model (ScanNet) & \textbf{101.1} & \textbf{16.2} & \textbf{58.7} & \textbf{61.8} & \textbf{58.4} & \textbf{58.0} & \textbf{17.7} \\
    \bottomrule
\end{tabular}
}
\label{tab:backbone_ablation}
\end{minipage}
\hfill
\begin{minipage}{0.32\linewidth} 
\centering
\includegraphics[width=0.9\linewidth]{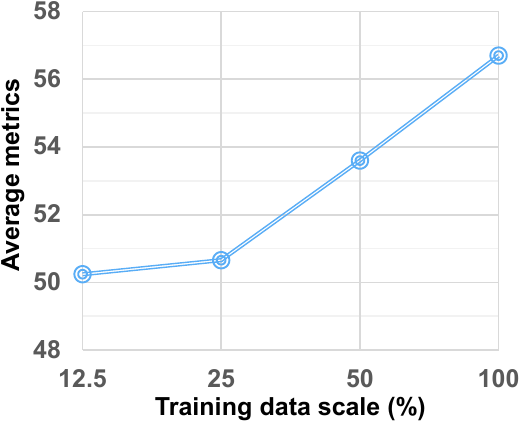}
\captionof{figure}{\textbf{Data scaling curve.} Performance is measured by the average metrics in \cref{tab:domain_ablation}.}
\label{fig:data_scaling}
\end{minipage}
\end{table*}

\begin{figure*}[t!]
\centering
\begin{minipage}{0.48\linewidth}
    \centering
    \includegraphics[width=\linewidth]{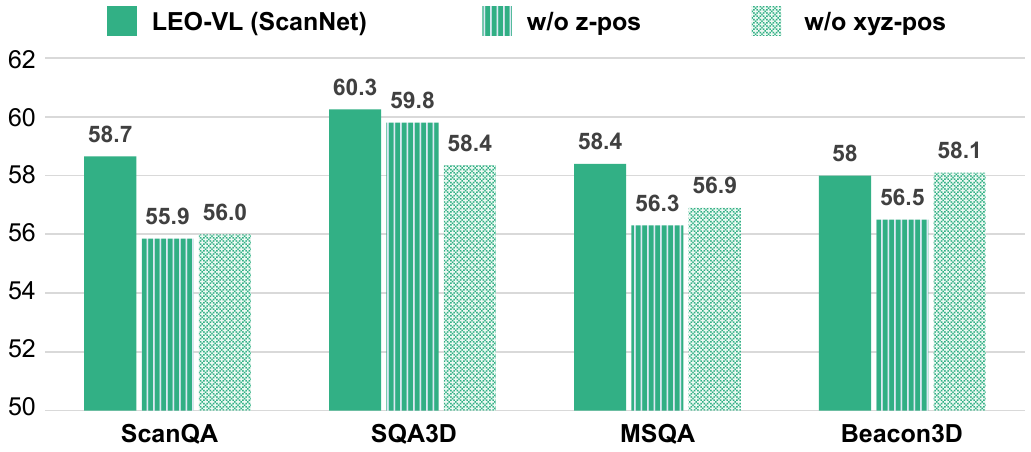}
    \captionof{figure}{\textbf{Ablation of position embeddings on the ScanNet subset.} For more consistent visualization, we use the case-centric metric for Beacon3D, and averaged metrics for others.}
    \label{fig:pos_ablation}
\end{minipage}
\hfill
\begin{minipage}{0.48\linewidth} 
    \centering
    \includegraphics[width=\linewidth]{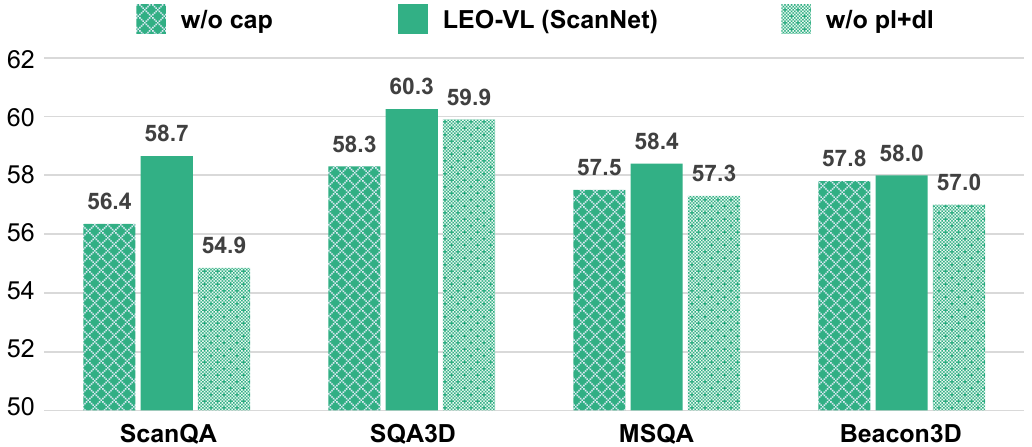}
    \captionof{figure}{\textbf{Ablation of task effects on the ScanNet subset.} For more consistent visualization, we use the case-centric metric for Beacon3D, and averaged metrics for others.}
    \label{fig:task_ablation}
\end{minipage}
\end{figure*}

\paragraph{3D Spatial Modeling.} A key design of \ac{cfg} lies in the disentanglement of 3D spatial modeling across vertical and horizontal directions. We train \model on the ScanNet subset under two ablated settings: ``\textbf{w/o z-pos}'', where the z-axis position embedding is removed; and ``\textbf{w/o xyz-pos}'', where both z-axis and xy-plane position embeddings are removed. As shown in \cref{fig:pos_ablation}, both types of position embeddings play critical roles in \ac{3dvl} tasks. Removing the z-axis position embedding leads to a general performance drop, while removing the xy-plane position embedding especially degrades performance on SQA3D. These results validate the effectiveness of our design in 3D spatial modeling.

\paragraph{Inference Efficiency.} We evaluate the inference efficiency of \model and compare to following baselines: LLaVA-3D \cite{zhu2024llava}, Video-3D LLM \cite{zheng2025video}, and Qwen2.5-VL-7B \cite{bai2025qwen2}, \ie, the backbone of \model without \ac{cfg}. Metrics are reported on ScanNet scene0050\_00 using the same prompt with an NVIDIA 4090 GPU. As shown in \cref{tab:inference_efficiency}, \model exhibits the highest efficiency, especially in FLOPs and time, highlighting the superior efficiency afforded by our \ac{cfg} representation.

\begin{table}[t!]
\captionof{table}{\textbf{Comparison of inference efficiency.} $^\text{\textdagger}$ indicates the backbone of \model without \ac{cfg}. All metrics are lower better.}
\vspace{0.2em}
\resizebox{\linewidth}{!}{
\begin{tabular}{lccc}
    \toprule
    Model & FLOPs (T) & Time (s) & Memory (GB) \\
    \midrule
    LLaVA-3D \cite{zhu2024llava} & 23.3 & 0.240 & 14.961 \\
    Video-3D LLM \cite{zheng2025video} & 116.7 & 1.375 & 21.689 \\
    Qwen2.5-VL-7B$^\text{\textdagger}$ \cite{bai2025qwen2} & 114.1 & 1.122 & 17.947 \\
    \model & \textbf{13.8} & \textbf{0.170} & \textbf{14.959} \\
    \bottomrule
\end{tabular}
}
\label{tab:inference_efficiency}
\end{table}

\subsection{Data Analysis}
\label{sec:data_ablation}

We conduct ablative studies on data to answer the following questions: (1) What is the effect of captioning, planning and dialogue tasks in \ac{3dvl} learning? (2) What is the benefit of scaling across diverse scene domains? (3) What happens when we naively scale up \ac{3dvl} data without considering data quality? (4) How effective is data scaling for \model with curated data?

\paragraph{Task Effects.} Based on \model (ScanNet), we ablate the training data with different task configurations: ``\textbf{w/o cap}'' removes both object captioning and scene captioning tasks, while ``\textbf{w/o pl+dl}'' removes planning and dialogue tasks. As shown in \cref{fig:task_ablation}, both task categories contribute meaningfully to the capabilities of \model. Captioning proves particularly beneficial for SQA3D, whereas planning and dialogue tasks are more influential for ScanQA. We further qualitatively test the ``\textbf{w/o cap}'' model to perform the scene captioning task. The model struggles to generate meaningful scene descriptions, likely due to the lack of training in captioning. These findings underscore the importance of task diversity for both acquiring task-specific skills and enhancing cross-task performance.

\begin{figure*}[t!]
\centering
\begin{minipage}{0.63\linewidth}
\centering
\captionof{table}{\textbf{Ablation results of scene domain scaling and simplistic QA data.} Benchmarks metrics follow \cref{tab:overall_results,tab:msqa,tab:beacon3d}, with metrics averaged per benchmark.}
\resizebox{\linewidth}{!}{
\begin{tabular}{lccccccc}
    \toprule
     Data & \textcolor{scannet}{ScanQA} & \textcolor{scannet}{SQA3D} & \textcolor{scannet}{MSQA} & \textcolor{scannet}{Beacon3D} & \textcolor{3rscan}{MSQA} & \textcolor{arkitscenes}{MSQA} & Avg. \\
    \midrule
    \rowcolor[HTML]{F0F7FF}\multicolumn{8}{l}{\textit{Scene domain scaling}} \\
    \textcolor{scannet}{ScanNet} & \textbf{58.7} & 60.3 & 58.4 & 37.9 & 46.1 & 54.7 & 52.7 \\
    \textcolor{scannet}{Scan}+\textcolor{3rscan}{3R} & 56.5 & 59.8 & 56.0 & 37.5 & 51.0 & 56.1 & 52.8 \\
    All & 58.0 & \textbf{62.3} & \textbf{61.7} & \textbf{39.4} & \textbf{52.0} & \textbf{66.5} & \textbf{56.7} \\
    \midrule
    \rowcolor[HTML]{F0F7FF}\multicolumn{8}{l}{\textit{Impact of simplistic QA data}} \\
    \textcolor{scannet}{Scan}+\textcolor{3rscan}{3R} & \textbf{56.5} & \textbf{59.8} & 56.0 & \textbf{37.5} & 51.0 & \textbf{56.1} & \textbf{52.8} \\
    + $\Delta$QA & 55.7 & 58.3 & \textbf{56.8} & 37.5 & \textbf{51.3} & 55.6 & 52.5 \\
    \bottomrule
\end{tabular}
}
\label{tab:domain_ablation}
\end{minipage}
\hfill
\begin{minipage}{0.35\linewidth} 
\centering
\captionof{table}{\textbf{QA data statistics.} ``$\Delta$QA'' refers to the extra part of QA data. ``Top-15 occupancy'' denotes the occupancy of the top 15 most frequent answer templates, measuring the diversity of QA data distribution. Higher top-15 occupancy indicates less diverse (more simplistic) data.}
\vspace{0.2em}
\resizebox{\linewidth}{!}{
\begin{tabular}{lcc}
    \toprule
     & Data scale & Top-15 occupancy \\
    \midrule
    \textcolor{scannet}{Scan}+\textcolor{3rscan}{3R} & 253k & 12.8\% \\
    $\Delta$QA & 669k & 39.2\% \\
    \bottomrule
\end{tabular}
}
\label{tab:simple_qa_stats}
\end{minipage}
\end{figure*}

\paragraph{Scene Domains.} To assess the impact of scaling across scene domains, we add a checkpoint trained on ScanNet and 3RScan (``\textcolor{scannet}{Scan}+\textcolor{3rscan}{3R}''). As shown in \cref{tab:domain_ablation}, ``\textcolor{scannet}{Scan}+\textcolor{3rscan}{3R}'' yields significant improvements on \textcolor{3rscan}{3RScan} and \textcolor{arkitscenes}{ARKitScenes} compared to ``\textcolor{scannet}{ScanNet}''. Furthermore, ``All'' shows consistent improvements on all domains compared to ``\textcolor{scannet}{Scan}+\textcolor{3rscan}{3R}''. These results suggest that the performance gains stem from the training in more diverse domains, emphasizing the importance of cross-domain scaling for \ac{3dvl} learning.

\paragraph{Priority of Quality over Scale.} Based on the results of ``\textcolor{scannet}{Scan}+\textcolor{3rscan}{3R}'', we explore the outcome of naively scaling up \ac{3dvl} data without considering data quality. We expand our default subsets on ScanNet and 3RScan with an extra 669k 3D QA data (denoted as ``$\Delta$QA'') from two sources: (1) LEO \cite{huang2024embodied}, which provides 94k QA samples from 3RScan; and (2) MMScan \cite{lyu2024mmscan}, which offers 575k QA samples from ScanNet and 3RScan. To measure the distribution of QA data, we categorize answers into templates and compute the proportion of the top 15 most frequent templates (top-15 occupancy). As shown in \cref{tab:simple_qa_stats}, ``$\Delta$QA'' comprises much more data than the default ``\textcolor{scannet}{Scan}+\textcolor{3rscan}{3R}'', and exhibits more simplistic data distribution. The results in \cref{tab:domain_ablation} reveal that despite substantially more QA data, ``+ $\Delta$QA'' shows degradation in the overall performance compared to ``\textcolor{scannet}{Scan}+\textcolor{3rscan}{3R}''. This shows the harm of scaling with low-quality data, and suggests that prioritizing data quality over scale is critical for effective scaling of \ac{3dvl} learning.

\paragraph{Consistent Scaling Effects.} We demonstrate that our curated data scheme exhibits consistent scaling effects. Specifically, we take 12.5\%, 25\%, and 50\% of our full data for training, and report the overall performance by averaging metrics across all benchmarks (following \cref{tab:domain_ablation}). As shown in \cref{fig:data_scaling}, the results exhibit a steady upward trend, demonstrating consistent gains as the data scale increases. This reflects the desired scaling behavior enabled by our curated data scheme, in contrast to the degradation when scaling with low-quality data. Notably, the upward trend shows no sign of saturation, suggesting the potential of further scaling with curated \ac{3dvl} data.

\begin{table}[t!]
\centering
\captionof{table}{\textbf{Evaluation results of post-training on SQA3D.} Baseline denotes the checkpoint before post-training. \textcolor{scannet}{SQA3D} represents \acf{id} evaluation, while \textcolor{3rscan}{Beacon3D} and \textcolor{multiscan}{Beacon3D} indicate \acf{ood} generalization.}
\resizebox{\linewidth}{!}{
\begin{tabular}{lcccc}
    \toprule
    \multirow{2}{*}{\raisebox{-0.8ex}{Post-training}} & \multicolumn{2}{c}{\textcolor{scannet}{SQA3D}} & \textcolor{3rscan}{Beacon3D} & \textcolor{multiscan}{Beacon3D} \\
    \cmidrule(lr){2-3} \cmidrule(lr){4-4} \cmidrule(lr){5-5}
    & EM & EM-R & EM-R & EM-R \\
    \midrule
    Baseline & 59.7 & 62.6 & 47.7 & 36.7 \\
    \midrule
    \ac{sft} & 61.0 & 64.0 & 48.8 & 36.7 \\
    \ac{grpo} & 39.9 & 63.3 & \textbf{50.0} & \textbf{43.4} \\
    SceneDPO & \textbf{61.1} & \textbf{64.2} & \textbf{50.0} & 40.3 \\
    \bottomrule
\end{tabular}
}
\label{tab:dpo_results}
\end{table}

\begin{figure}[t!] 
\centering
\includegraphics[width=\linewidth]{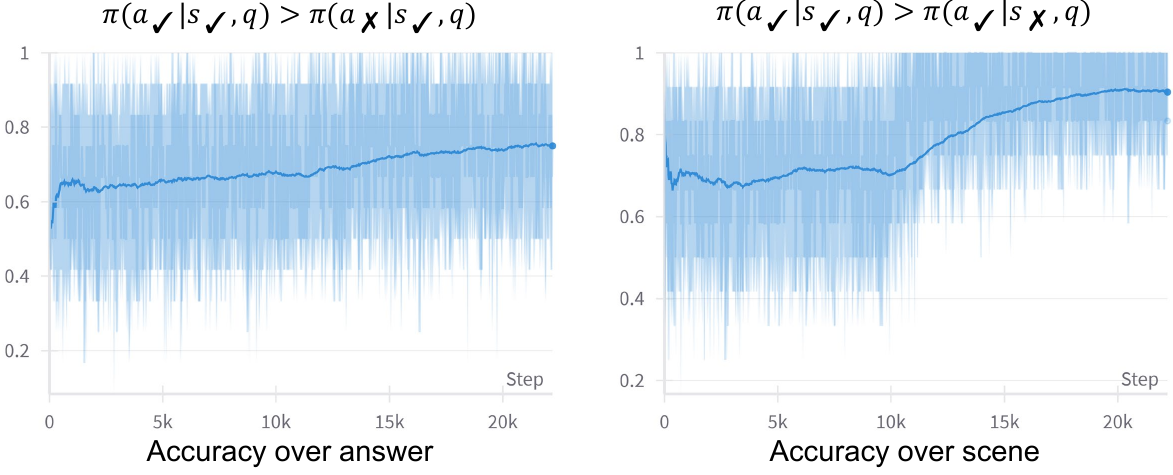}
\captionof{figure}{\textbf{Dynamics of contrastive accuracy during post-training.}}
\label{fig:dpo_curves}
\end{figure}

\subsection{Post-training}
\label{sec:post_training}

\paragraph{Settings.} To explore post-training strategies for 3D \acp{vlm}, we continue to train \model on SQA3D using the SceneDPO objective, comparing it against \ac{sft} and \ac{grpo}. Evaluation encompasses \acf{id} tests on SQA3D (\textcolor{scannet}{ScanNet}) and \acf{ood} tests on Beacon3D (\textcolor{3rscan}{3RScan} and \textcolor{multiscan}{MultiScan}), utilizing EM and EM-R as primary metrics. Training hyperparameters are consistent with the instruction-tuning stage. Specific implementations are detailed below:
\begin{itemize}[nolistsep,noitemsep,leftmargin=*]
    \item \textbf{\ac{sft}.} We follow the previous instruction-tuning stage.
    \item \textbf{\ac{grpo}.} We design an accuracy reward that yields $1$ for EM correct, $0.2$ for EM-R correct, and $0$ otherwise. Format reward is omitted due to the absence of \ac{cot} data. We set the group size to $4$, clip range $\epsilon=0.2$, and KL coefficient $\beta=0.1$.
    \item \textbf{SceneDPO.} Negative answers ($a_\text{\xmark}$) are bootstrapped from model predictions by retaining incorrect outputs or rephrasing correct ones into distractors. Negative scenes ($s_\text{\xmark}$) are randomly sampled from the other scenes as long as $s_\text{\xmark} \neq s_\text{\cmark}$. We set $w_a=0.5$, $w_s=0.5$, $\beta_a=0.2$, and $\beta_s=0.03$.
\end{itemize}

\paragraph{Evaluation Results.} As shown in \cref{tab:dpo_results}, while most post-training strategies improve upon the baseline, GRPO struggles on SQA3D. This implies the GRPO's weakness in enhancing \ac{id} task, probably because the reward signals are too sparse to facilitate complex spatial reasoning in 3D \acp{vlm}. However, GRPO exhibits better \ac{ood} performance than SFT, indicating its comparative advantage in generalizability. In contrast to \ac{grpo}'s weak \ac{id} performance and \ac{sft}'s weak \ac{ood} performance, SceneDPO achieves both strong \ac{id} and \ac{ood} performances, addressing the limitations of both \ac{sft} and \ac{grpo}. These findings position SceneDPO as a more practical and robust post-training objective for 3D \acp{vlm}.

\paragraph{Training Dynamics.} \cref{fig:dpo_curves} visualizes the dynamic curves of contrastive accuracy for answers and scenes during post-training. We observe a consistent upward trend in both, reflecting the model's increasing ability to discriminate between positive and negative pairs. Notably, the initially low scene-contrast accuracy reveals the underlying ``visual ignorance'' issue, \ie, a form of hallucination where the model fails to exploit scene context for QA. The subsequent improvement demonstrates that SceneDPO effectively mitigates this phenomenon, enhancing the model's grounded reasoning. These observations resonate with our design motivation and demonstrate the efficacy of SceneDPO for enhancing the robustness of 3D \acp{vlm}.

\paragraph{Loss Ablation.} Ablation results in \cref{tab:scenedpo_ablation} confirm the necessity of both NLL loss ($\mathcal{L}_\text{NLL}$) and scene-contrast loss ($\mathcal{L}_s$). Specifically, scene-contrast loss ($\mathcal{L}_s$) drives significant gains in \ac{ood} generalization ($50.4\rightarrow52.8$ on \textcolor{3rscan}{Beacon3D}; $37.7\rightarrow39.4$ on \textcolor{multiscan}{Beacon3D}), while NLL loss ($\mathcal{L}_\text{NLL}$) prevents degradation in \ac{id} performance ($59.6\rightarrow61.1$ on \textcolor{scannet}{SQA3D} EM; $62.6\rightarrow64.2$ on \textcolor{scannet}{SQA3D} EM-R). These results underscore that SceneDPO successfully adapts the DPO framework to the context of \ac{3dvl} post-training, effectively remedying the limitations of vanilla DPO to provide a robust post-training objective for 3D \acp{vlm}.

\begin{figure*}[t!]
    \centering
    \includegraphics[width=\linewidth]{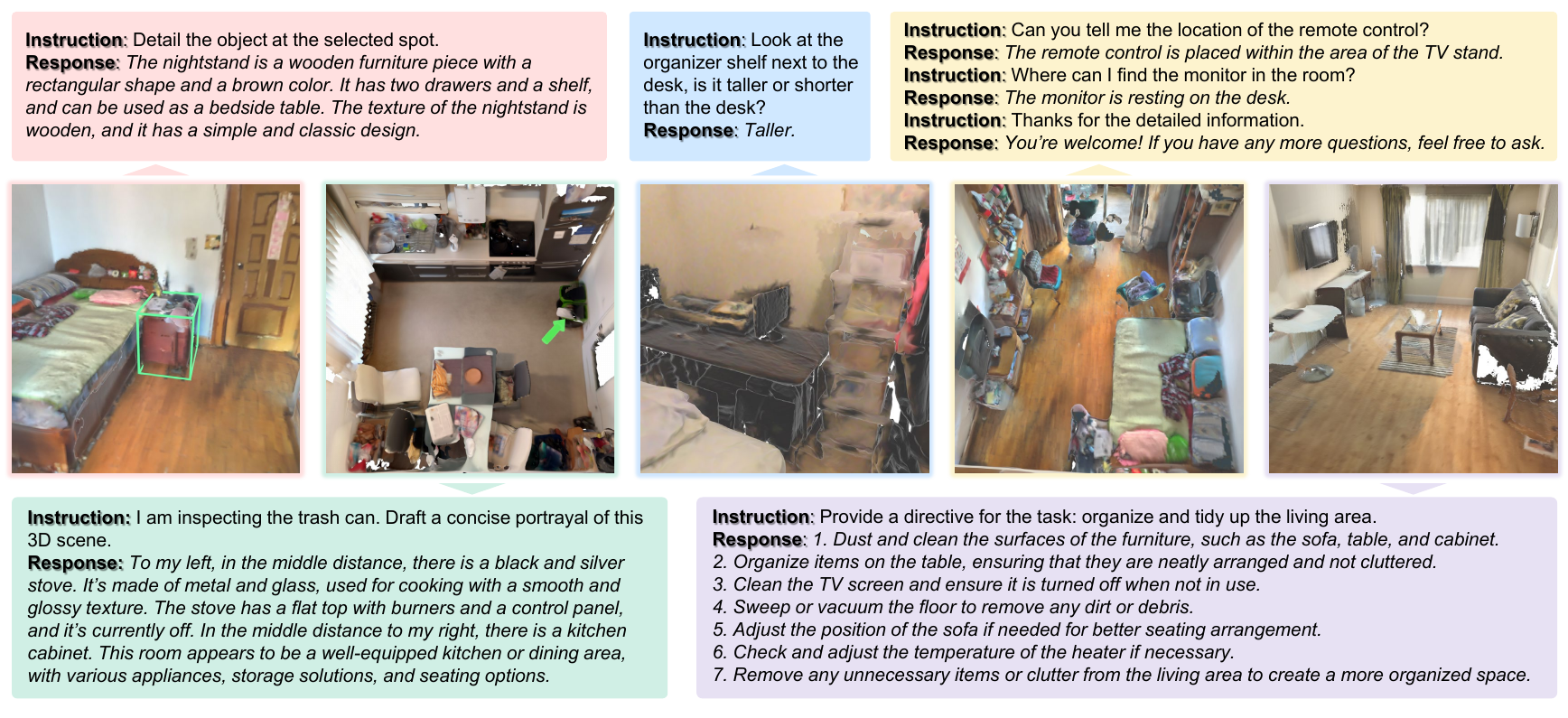}
    \captionof{figure}{\textbf{Qualitative results of \model on various tasks.} The top row illustrates object captioning (red), question answering (blue), and dialogue (yellow). The bottom row illustrates scene captioning (green) and planning (purple).}
    \label{fig:qualitative}
\end{figure*}

\begin{table*}[t!]
\centering
\small
\vspace{0.4em}
\captionof{table}{\textbf{Failure cases on the vertical-spatial-relation subset.} Questions encompass ``What is (directly) on/above/under/below ...''}
\vspace{0.2em}
\resizebox{\linewidth}{!}{
\begin{tabular}{lcp{5.5cm}p{6.5cm}}
    \toprule
    Failure Category & Count & Definition & Example \\
    \midrule
    Hallucination/Overfitting& 7/15 & The model answers with a hallucinated object based on prior knowledge & Q: What is on the brown ottoman? \newline A: Laptop (actually there is no laptop) \\
    \midrule
    Confusion on Overlapped Objects& 3/15 & The model mistakes objects that are close within the same pillar area & Q: What is above the large bed? \newline A: Pillow (GT: Picture, while pillow is on the bed) \\
    \midrule
    Grounding/Classification Failure& 5/15 & The model fails to ground or classify the related object instance & Q: What is under ``the stove with pots on it''? \newline A: Dishwasher (GT: Oven, looks like dishwasher) \\
    \bottomrule
\end{tabular}
}
\label{tab:failure_case}
\end{table*}

\begin{table}[t!]
\centering
\captionof{table}{\textbf{SceneDPO loss ablation on SQA3D post-training.} \textcolor{scannet}{SQA3D} represents \acf{id} evaluation, while \textcolor{3rscan}{Beacon3D} and \textcolor{multiscan}{Beacon3D} indicate \acf{ood} generalization.}
\vspace{0.2em}
\resizebox{\linewidth}{!}{
\begin{tabular}{lcccc}
    \toprule
    \multirow{2}{*}{\raisebox{-0.8ex}{Loss}} & \multicolumn{2}{c}{\textcolor{scannet}{SQA3D}} & \textcolor{3rscan}{Beacon3D} & \textcolor{multiscan}{Beacon3D} \\
    \cmidrule(lr){2-3} \cmidrule(lr){4-4} \cmidrule(lr){5-5}
    & EM & EM-R & EM-R & EM-R \\
    \midrule
    SceneDPO & \textbf{61.1} & \textbf{64.2} & 50.0 & \textbf{40.3} \\
    - NLL ($\mathcal{L}_\text{NLL}$) & 59.6 & 62.6 & \textbf{52.8} & 39.4 \\
    - scene ($\mathcal{L}_s$) & 60.3 & 63.2 & 50.4 & 37.7 \\
    \bottomrule
\end{tabular}
}
\label{tab:scenedpo_ablation}
\end{table}

\subsection{Case Analysis}
\paragraph{Vertical Spatial Relations.} To probe \model's ability to handle vertical spatial relations, we curate a specialized ``vertical-spatial-relation'' subset of 34 questions from the \textcolor{scannet}{Beacon3D} dataset. These questions encompass patterns like ``What is (directly) on/above/under/below ...'' \model achieves an accuracy of 58.1\% on this subset, which is notably higher than the 41.2\% accuracy on the broader parent category. This suggests that vertical condensation does not introduce a specific bottleneck for vertical spatial reasoning. Furthermore, we identify and categorize 15 failure cases within this subset, as detailed in \cref{tab:failure_case}. We observe that the primary failure mode remains general hallucination/overfitting, rather than confusion regarding overlapped objects. These observations confirm that our architectural design successfully leverages \ac{cfg} to improve representation efficiency without compromising the model's ability to reason about 3D spatial structures.

\subsection{Qualitative Examples}

We present qualitative results of \model performing various tasks in \cref{fig:qualitative}, including object captioning, scene captioning, \acf{qa}, planning, and dialogue. These examples cover 3D scenes from ScanNet, 3RScan, MultiScan, and ARKitScenes, illustrating the versatile capabilities of \model in diverse 3D scenes.

\vspace{1em}

\section{Conclusion}
\label{sec:conclusion}

We introduce \model, an efficient 3D \ac{vlm} featuring versatile capabilities across diverse 3D scene domains. At the core of our approach is the \acf{cfg}, a scene representation that significantly reduces token overhead while preserving global 3D spatial structures and simultaneously simplifying spatial modeling. This optimizes the capability-efficiency Pareto frontier and unlocks the scalability of \ac{3dvl} learning. Consequently, we curate a comprehensive \ac{3dvl} dataset spanning four real-world indoor domains (ScanNet, 3RScan, MultiScan, and ARKitScenes) and five tasks, including captioning, \acf{qa}, planning, and dialogue. Experimental results demonstrate that \model achieves \sota performance across various \ac{3dvl} benchmarks such as SQA3D, Beacon3D, and Scan2Cap. Our extensive analyses provide valuable insights, such as the importance of diversity in tasks and scene domains, and the necessity of data curation for effective scaling. Furthermore, we propose SceneDPO, a robust post-training objective for 3D \acp{vlm}, and demonstrate its advantages compared to \ac{sft} and \ac{grpo}. We hope our findings advance the development of efficient, scalable, and robust 3D \acp{vlm}.

\paragraph{Limitations.}
Despite strong performance, \model has several limitations. First, the \ac{cfg} representation may be insufficient for tasks requiring high spatial precision, such as fine-grained localization. Second, as our data primarily covers indoor scenes, generalization to outdoor or dynamic environments remains underexplored. Third, the efficacy of SceneDPO depends on the quality of contrastive pairs. Future directions include more expressive representations, broader scene coverage, and streamlined post-training strategies.

\bibliography{ref}
\bibliographystyle{IEEEtran}

\end{document}